\documentclass[sigconf,anonymous=false]{acmart}

\usepackage{multirow}
\usepackage{pifont}% http://ctan.org/pkg/pifont
\usepackage[utf8]{inputenc}  
\usepackage{svg}
\usepackage{tikz}  
\usepackage{pgfplots}
\usepackage{algorithm}
\usepackage{algpseudocode}
\usepackage{dsfont}
\usepackage{subcaption}
\usepackage{multirow}  
\def\({{\big (}}
\def\){\big )}

\makeatletter
\newcommand*{\bigcdot}{}% Check if undefined
\DeclareRobustCommand*{\bigcdot}{%
  \mathbin{\mathpalette\bigcdot@{}}%
}
\newcommand*{\bigcdot@scalefactor}{.5}
\newcommand*{\bigcdot@widthfactor}{1.15}
\newcommand*{\bigcdot@}[2]{%
  \sbox0{$#1\vcenter{}$}% math axis
  \sbox2{$#1\cdot\m@th$}%
  \hbox to \bigcdot@widthfactor\wd2{%
    \hfil
    \raise\ht0\hbox{%
      \scalebox{\bigcdot@scalefactor}{%
        \lower\ht0\hbox{$#1\bullet\m@th$}%
      }%
    }%
    \hfil
  }%
}
\makeatother

\makeatletter

\def\env@sqcases{%https://www.overleaf.com/project/5f5a9a85c3bfc50001334c55
  \let\@ifnextchar\new@ifnextchar
  \left\lbrack
  \def\arraystretch{1.2}%
  \array{@{}l@{\quad}l@{}}%
}
\makeatother
\AtBeginDocument{%
  }

\setcopyright{acmlicensed}
\copyrightyear{2025}
\acmYear{2025}
\acmDOI{XXXXXXX.XXXXXXX}

\acmConference[KDD '25]{31st ACM SIGKDD Conference on Knowledge Discovery and Data Mining}{3rd–7th August 2025}{Toronto, Canada}
\acmISBN{978-1-4503-XXXX-X/18/06}

\begin{document}

% \title{Multi-Epoch Large Scale Conversion Modeling Using Sparse Embeddings with Frequency-Adaptive Learning Rate}
\title{The Evolution of Embedding Table Optimization and Multi-Epoch Training in Pinterest Ads Conversion}

\author{Andrew Qiu}
\affiliation{
  \institution{Pinterest Inc.}
  \streetaddress{85 Richmond St W 8th floor}
  \city{Toronto}
  \state{Ontario}
  \country{Canada}
}
\email{aqiu@pinterest.com}

\author{Shubham Barhate}
\affiliation{
  \institution{Pinterest Inc.}
  \streetaddress{85 Richmond St W 8th floor}
  \city{Toronto}
  \state{Ontario}
  \country{Canada}
}
\email{shubhambarhate@pinterest.com}

\author{Hin Wai Lui}
\affiliation{
  \institution{University of California Irvine}
  \streetaddress{Irvine, CA 92697, USA}
  \city{Irvine}
  \state{California}
  \country{USA}
}
\email{<get email from Runze>}

\author{Runze Su}
\affiliation{
  \institution{Pinterest Inc.}
  \streetaddress{651 Brannan St}
  \city{San Francisco}
  \state{California}
  \country{USA}
}
\email{runzesu@pinterest.com}

\author{Rafael Rios Müller}
\affiliation{
  \institution{Pinterest Inc.}
  \streetaddress{85 Richmond St W 8th floor}
  \city{Toronto}
  \state{Ontario}
  \country{Canada}
}
\email{rmuller@pinterest.com}

\author{Kungang Li}
\affiliation{
  \institution{Pinterest Inc.}
  \streetaddress{651 Brannan St}
  \city{San Francisco}
  \state{California}
  \country{USA}
}
\email{kungangli@pinterest.com}

\author{Ling Leng}
\affiliation{
  \institution{Pinterest Inc.}
  \streetaddress{651 Brannan St}
  \city{San Francisco}
  \state{California}
  \country{USA}
}
\email{lleng@pinterest.com}

\author{Han Sun}
\affiliation{
  \institution{Pinterest Inc.}
  \streetaddress{651 Brannan St}
  \city{San Francisco}
  \state{California}
  \country{USA}
}
\email{hsun@pinterest.com}

\author{Shayan Ehsani}
\affiliation{
  \institution{Pinterest Inc.}
  \streetaddress{85 Richmond St W 8th floor}
  \city{Toronto}
  \state{Ontario}
  \country{Canada}
}
\email{sehsani@pinterest.com}

\author{Zhifang Liu}
\affiliation{
  \institution{Pinterest Inc.}
  \streetaddress{651 Brannan St}
  \city{San Francisco}
  \state{California}
  \country{USA}
}
\email{zhifangliu@pinterest.com}

%%\email{aqiu, shubhambarhate, runzesu, rmuller, hlui}

%%\email{sehsani, hsun, kungangli, lleng @pinterest.com}


\begin{abstract}

Deep learning for conversion prediction has found widespread applications in online advertising. These models have become more complex as they are trained to jointly predict multiple objectives such as click, add-to-cart, checkout and other conversion types. Additionally, the capacity and performance of these models can often be increased with the use of embedding tables that encode high cardinality categorical features such as advertiser, user, campaign, and product identifiers (IDs). These embedding tables can be pre-trained, but also learned end-to-end jointly with the model to directly optimize the model objectives. Training these large tables is challenging due to: gradient sparsity, the high cardinality of the categorical features, the non-uniform distribution of IDs and the very high label sparsity. These issues make training prone to both slow  convergence and overfitting after the first epoch. Previous works addressed the multi-epoch overfitting issue by using: stronger feature hashing to reduce cardinality, filtering of low frequency IDs, regularization of the embedding tables, re-initialization of the embedding tables after each epoch, etc. Some of these techniques reduce overfitting at the expense of reduced model performance if used too aggressively. In this paper, we share key learnings from the development of embedding table optimization and multi-epoch training in Pinterest Ads Conversion models. We showcase how our Sparse Optimizer speeds up convergence, and how multi-epoch overfitting varies in severity between different objectives in a multi-task model depending on label sparsity. We propose a new approach to deal with multi-epoch overfitting: the use of a frequency-adaptive learning rate on the embedding tables and compare it to embedding re-initialization. We evaluate both methods offline using an industrial large-scale production dataset.

\end{abstract}


\begin{CCSXML}
<ccs2012>
<concept>
<concept_id>10002951</concept_id>
<concept_desc>Information systems</concept_desc>
<concept_significance>500</concept_significance>
</concept>
<concept>
<concept_id>10002951.10003227</concept_id>
<concept_desc>Information systems~Information systems applications</concept_desc>
<concept_significance>500</concept_significance>
</concept>
<concept>
<concept_id>10002951.10003227.10003447</concept_id>
<concept_desc>Information systems~Computational advertising</concept_desc>
<concept_significance>500</concept_significance>
</concept>
<concept>
<concept_id>10002951.10003227.10003233.10010519</concept_id>
<concept_desc>Information systems~Social networking sites</concept_desc>
<concept_significance>300</concept_significance>
</concept>
</ccs2012>
\end{CCSXML}

\ccsdesc[500]{Information systems}
\ccsdesc[500]{Information systems~Information systems applications}
\ccsdesc[500]{Information systems~Computational advertising}
\ccsdesc[300]{Information systems~Social networking sites}

\keywords{Ads Recommendation Systems, Click-Through Rate Prediction, Training Optimization, Overfitting, Multi-Epoch Learning}
%   \includegraphics[width=\textwidth]{figures/oCPM_Model_Architecture_Figure.pdf}
%   \caption{oCPM Multi-Task Model Architecture}
%   \label{fig:architecture}

\maketitle
%/////////////////////////////////////////////////////////////////////////////////////
\section{Introduction}\label{sec:introduction}
%/////////////////////////////////////////////////////////////////////////////////////
The complexity of deep learning models for conversion prediction in online advertising has spurred a rich body of research exploring deep hierarchical ensemble networks with different feature crossing modules, large embedding table scaling and more recently, multi-task learning setups that jointly train related target objectives such as click and conversion prediction \cite{zhang2022dhendeephierarchicalensemble, guo2024embeddingcollapsescalingrecommendation, wang2023multitaskdeeprecommendersystems}.

A critical component of these models is the use of embedding tables, which play a central role by encoding high-cardinality categorical features like advertiser and product identifiers. However, training these tables presents significant challenges. Since only a small portion of rows are updated within each batch, embedding tables experience slow convergence, as individual rows receive limited parameter updates during training \cite{li2021frequencyawaresgdefficientembedding}. Additionally, the high cardinality, non-uniform ID distribution, and extreme label sparsity contribute to a phenomenon known as the one-epoch phenomenon or multi-epoch overfitting. This phenomenon is characterized by sharp, unrecoverable drops in test performance at the start of new epochs. \cite{zhang2022towards, fan2024multi}. Training these models for more than one epoch is a desired property in industrial use-cases, since it can drive improved performance while also enabling the reduction of the data storage requirements by replicating the performance of a large dataset with multiple passes over a smaller dataset. 

Previous studies address multi-epoch overfitting through cardinality reduction techniques like the hashing trick or filtering of low frequency IDs \cite{zhang2022towards}. However, these methods sometimes compromise model performance by diluting the granularity of the embeddings. Alternative strategies involve global approaches across entire embedding tables, such as decreased learning rates, embedding re-initialization, and the application of various regularization techniques \cite{zhang2022towards, fan2024multi, wang2021dcn}. Despite partially mitigating overfitting, these methods often do not differentiate between infrequent and frequent rows within the table, potentially leaving room for optimization.

% Noteworthy advancements include multi-epoch learning frameworks and data augmentation methods designed to maintain embedding relative positions and enhance robustness, outperforming conventional single-epoch training in CTR models. Other contributions focus on managing high cardinality through adaptive frameworks that preserve cardinality while targeting overfitting challenges, offering a nuanced balance between accuracy and generalization.

In this paper, we share key learnings from the development of embedding table optimization and multi-epoch training in Pinterest Ads Conversion models. We introduce our Sparse Optimizer, which applies a higher layer-specific learning rate to embedding tables to both improve convergence speed and achieve better performance. Additionally, we discuss our model's adaptation to multi-epoch learning, highlighting the challenge of multi-epoch overfitting in a multi-task learning context. To address this, we propose a novel method called Frequency-Adaptive Learning Rate (FAL), which selectively reduces the learning rate for infrequent rows to mitigate multi-epoch overfitting while preserving the performance of frequently accessed rows. By integrating these methods, we achieve faster convergence of embedding tables and reduce performance loss from multi-epoch overfitting in most cases. We compare our method to an existing multi-epoch overfitting approach, embedding re-initializing (MEDA) \cite{fan2024multi}, on a large-scale industrial dataset and achieve comparable results on all objectives except the sparsest one. Although both MEDA and FAL achieve superior performance compared to a multi-epoch baseline on a validation set, these gains vanish after several days of continual training as the baseline catches up in performance. As such, we demonstrate how solving multi-epoch overfitting may be unnecessary as long as fresh data is available and overfitting is not too severe.

%/////////////////////////////////////////////////////////////////////////////////////
\section{Related Work}\label{sec:related_work}
\subsection{Embedding Table in DLRM}
Deep learning has become a cornerstone of online recommendation systems, driving improved user engagement and personalization \cite{zhang2021deep, gao2023rec4ad}. In extensive research on deep learning recommendation systems \cite{zhou2018deep, song2019autoint, mao2023finalmlp, wang2021dcn, zhang2023fibinet++, wang2021masknet}, embedding tables play a crucial role in their architecture and performance.

Early implementations of embedding tables sought to transform sparse categorical data into dense vector representations, significantly improving computational efficiency over previous methods such as One-Hot-Encoding \cite{pan2024ads} and Skip-gram \cite{covington2016deep, grbovic2018real, zhao2018learning}. 

\subsubsection{Cardinality Optimization}

While general hashing with skip-gram preserves semantic relationships through context windows, hash embedding \cite{weinberger2009feature} focuses on efficient memory usage by compressing high-cardinality datasets into reduced-size spaces, often trading detailed relational semantics for scalability and computational efficiency. The Multi-Hash technique \cite{tito2017hash} enhances traditional hashing by utilizing multiple hash functions with trainable parameters to reduce collisions and improve vector uniqueness. Hybrid Hashing \cite{zhang2020model} further refines this approach by considering feature frequency, using unique IDs for high-frequency features and double hashing for low-frequency ones to balance accuracy and efficiency. The QR Trick \cite{shi2020compositional} employs quotient and remainder operations to create unique embedding vectors, ensuring collision-free partitioning in high-cardinality data scenarios. Binary hash embedding \cite{yan2021binary} offers an innovative method that combines flexibility in adjusting the storage size of embeddings with maximizing model efficacy. 

\subsubsection{Dimension Optimization}

FITTED \cite{luo2024finegrainedembeddingdimensionoptimization} adaptively reduces both the cardinality and dimension of chunks of embedding rows based on frequency and gradient sizes. The overparameterization of embedding tables leaves it prone to neural or representation collapse, where the variability of cross-example activations is limited \cite{Papyan_2020, tirer2022extendedunconstrainedfeaturesmodel}. A similar effect is observed when scaling the embedding dimension, where high dimensional embedding tables tend to be low-rank \cite{guo2024embeddingcollapsescalingrecommendation}. The ensembling of disjoint embedding and feature interaction networks has been shown to improve embedding dimension scalability \cite{guo2024embeddingcollapsescalingrecommendation}.

\subsubsection{Slow Convergence}

Adaptive algorithms have been shown to increase the convergence speed of rarely seen features by scheduling a higher learning rate \cite{JMLR:v12:duchi11a}. Frequency-Adaptive SGD (FA-SGD) provably and experimentally shows faster convergence compared to both SGD and AdaGrad by adaptively assigning a higher learning rate to more infrequent rows \cite{li2021frequencyawaresgdefficientembedding} . 

\subsection{Multi-Epoch Overfitting}

High cardinality in hash embedding tables often results in increased sparsity\cite{guo2023embedding} and exacerbates cold-start issues \cite{xu2022alleviating}. A significant challenge in this area is multi-epoch overfitting, where models are prone to sharp, unrecoverable drops to test performance when trained beyond the first epoch \cite{zhang2022towards}. There are several methods to mitigating this overfitting. Regularization techniques \cite{zhang2020graph} constrain the embedding complexity, producing generalized representations that perform better on unseen data. Re-initializing the embedding table before the second epoch (MEDA) \cite{fan2024multi} can reset embedding representations and prevent the model from reinforcing non-generalizable patterns. Utilizing pre-trained models to initialize embedding tables \cite{hsu2024taming} can leverage existing knowledge and reduce the risk of early-phase overfitting. Implementing multiple embedding tables across various expert models \cite{pan2024ads, lin2024disentangled} also allows for diversified learning tasks and helps prevent concentrated overfitting on singular embedding spaces.

%/////////////////////////////////////////////////////////////////////////////////////

%/////////////////////////////////////////////////////////////////////////////////////
\section{Methodology}\label{sec:methodology}

\subsection{Sparse Optimizer}

Only a small subset of embedding table rows are updated in each batch, resulting in high gradient sparsity. Over the course of training, individual rows receive limited parameter updates leading to the slow convergence of embedding tables. To address this issue, we assign a higher layer-specific learning rate for embedding tables on top of our Adam optimizer \cite{kingma2017adammethodstochasticoptimization}. That is, we increase the learning rate of embedding table parameters while fixing the learning rate of all other (dense) parameters. At Pinterest, we refer to this approach as the Sparse Optimizer.

\subsection{Frequency-Adaptive Learning Rate (FAL)}

Multi-epoch overfitting is caused by the overfitting of infrequent rows in the embedding layer, followed by the rapid adaptation of downstream layers to these biases \cite{fan2024multi, zhang2022towards}. It has been shown that a lowered learning rate reduces multi-epoch overfitting at the cost of decreased peak performance \cite{zhang2022towards}, which may be a result of slower convergence and tendency to stay close to the small weights at initialization.

To prevent performance degradation for frequent IDs, we propose a frequency-adaptive learning rate (FAL) which assigns progressively lower learning rates to less frequent rows. FAL aggressively slows convergence for infrequent rows while minimally affecting frequent rows. We scale row-wise learning rates by their relative log frequency, as to prevent highly frequent rows from dominating the learning rate distribution. As seen in Table \ref{tab:lin_vs_log}, log scaling achieves higher offline performance in comparison to the linear scaling of relative frequency.

\begin{table}[h]
    \centering
        \caption{Comparative cumulative AUC gain on checkout | click over 10 days of continual training, after 60 days of 2-epoch training.}
    \begin{tabular}{l|c}
        \textbf{Scaling Method} & \textbf{Cumulative AUC gain} \\
        \hline
        Log Scaling (Control) & 0.00\%  \\
        Linear Scaling & -0.13\% \\
    \end{tabular}
    % https://reports.pinadmin.com/execution/M10N/Ranking/oCPM/MLEnv%20Offline%20Comparison/2025-02-06%2016:23:08

    % https://reports.pinadmin.com/execution/M10N/Ranking/oCPM/MLEnv%20Offline%20Comparison/2025-04-07%2018:27:29
    \label{tab:lin_vs_log}
\end{table}

In comparison to existing sparsity-aware optimization algorithms like Adagrad \cite{JMLR:v12:duchi11a} and FA-SGD \cite{li2021frequencyawaresgdefficientembedding}, FAL reduces the learning rate for less frequently updated parameters rather than increase them. Although these algorithms exhibit faster convergence, we identify that this could come at the cost of overfitting, as low frequency IDs disproportionately cause multi-epoch overfitting and higher learning rates exacerbate the issue \cite{zhang2022towards}. The balance between fast convergence and low overfitting is delicate. We believe that a combination of our Sparse Optimizer, which increases convergence for high or medium-frequency IDs which are less prone to overfit, and FAD, which punishes convergence speed for especially low-frequency IDs, may be a promising direction forward.

\subsubsection{Implementation}

Let $F_T[i] \in \mathbb{N}$ be the total frequency of the $i$th row of embedding table $T \in \mathbb{R}^{n \times d}$, with $n$ rows and embedding dimension $d$. Let $\eta_T^*$ be the layer-specific learning rate for $T$. We scale the row-specific learning rate $\eta_T[i]$ by their relative log frequencies:

\begin{equation}
     \eta_T[i] = \eta_T^* \cdot \frac{\log(F_T[i] + 1)}{\max_j \log(F_T[j] + 1)} 
\end{equation}

Let $f_T(B) \in \mathbb{N}^{n}$ be the embedding frequencies seen for table $T$ in batch $B$. To remove the need for precomputation, we accumulate frequencies over training as seen in Algorithm \ref{alg:FAL-impl}. In offline testing, we observe negligible performance difference with or without pre-computation. We store the frequencies as a 32-bit unsigned integer tensor per table. This results in negligible GPU memory overhead, equivalent to increasing the embedding dimension of each table by one. We use a fixed embedding dimension of 32, so FAL results in a memory overhead of $3.125\%$ for embedding tables, with no effect to other layers.

FAL affects the learning rate of individual embedding rows. We may interpret FAL as a per-embedding learning rate schedule. As such, FAL does not affect the use of gradient clipping, and momentum terms used in adaptive optimizers like Adam. 

\begin{algorithm}
    \caption{FAL Implementation}
    \label{alg:FAL-impl}
    \begin{algorithmic}
        \State $F_T \gets \mathbf{0} \in \mathbb{N}^{n}$ for each Embedding Table $T$
        \For{batch $B$}
            \State Forward Pass
            \State Backward Pass
            \For{Embedding Table $T$}
                \State $F_T \gets F_T + f_T(B)$
                \State $\alpha_T \gets \log(1 + F_T) / \max (\log (1 + F_T))$
                \State $\nabla T \gets \nabla T \cdot \alpha_T$
            \EndFor
            \State Parameter Update
        \EndFor
    \end{algorithmic}
\end{algorithm}

%/////////////////////////////////////////////////////////////////////////////////////

\section{Ads Conversion Modeling at Pinterest}

\subsection{Dataset}
The training dataset is constructed by joining the impression logs with on-site and off-site conversion actions up to an attribution window delay. The positive labels in the dataset are clicks, good clicks (clicks with a longer duration), add-to-carts, and checkouts. A breakdown of the relative frequencies of positive labels in our dataset is provided in Table 1. Negative examples are aggressively downsampled.

\begin{table}[h]  
    \centering  
    \caption{Label density of different objectives}  
    \begin{tabular}{l|c}  
        \textbf{Objective} & \begin{tabular}[c]{@{}c@{}}\textbf{Positive label density}\\ \textbf{relative to clicks}\end{tabular} \\  
        \hline  
        $p$(checkout | click) & 0.002 \\  
        $p$(checkout | view, no click) & 0.01 \\  
        $p$(add-to-cart | click) & 0.007 \\  
        $p$(add-to-cart | view, no click) & 0.03 \\  
        $p$(good click) & 0.15 \\  
        $p$(click) & 1.00 \\  
    \end{tabular}  
    \label{tab:head_density}  
\end{table}  

\begin{figure}[h]
  \centering
  \includegraphics[width=\linewidth]{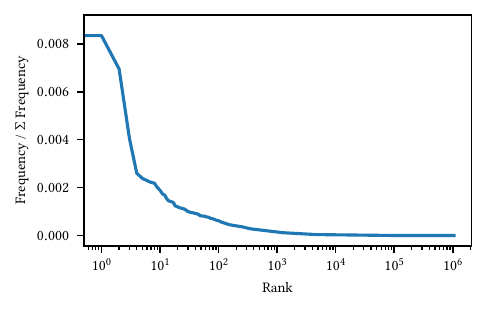}
  \caption{Frequency distribution of the categorical feature "Campaign ID" post-hashing.}
  \label{fig:campaign_freq_dist}
\end{figure}

Our dataset encompasses a diverse range of features, such as pre-trained user and ad pin embeddings, categorical and entity ID features, entity-level engagement metrics, user-related sequence attributes, and user-entity interaction count features. Consistent with findings from  \cite{zhang2022towards, fan2024multi}, our entity ID features exhibit highly non-uniform distributions. Specifically, a small subset of IDs occur with very high frequency, while most IDs appear infrequently. Figure \ref{fig:campaign_freq_dist} shows the frequency distribution of "Campaign ID", in which 50\% of the frequency is captured by 0.74\% of IDs.

% Our dataset can be optionally filtered into a shopping slice, which includes shoppable product pins that directly embed product information such as availability and pricing. This slice represents X\% of our dataset. In particular, as product IDs are only applicable to product pins, they exhibit greater sparsity than campaign and advertiser IDs, which are present for each ad. We expect the shopping slice to be particularly affected by multi-epoch overfitting due to its unique reliance on very sparse product ID features.

\subsection{Model Training Setup}

Our model is initially trained via a long duration "batch training" over a dataset encompassing 110 consecutive days. The order of the dataset is shuffled. After batch training, we continually train our model on single-day, fresh datasets. 

\subsection{Model Training Parameters}

We use the Adam Optimizer \cite{kingma2017adammethodstochasticoptimization}, with a base learning rate of 0.00015, batch size of 2000, $\epsilon=0.00001$, and $\beta=(0.9, 0.999)$.

\subsection{Offline Evaluation Metrics}

After each training, we evaluate the model using unseen, next-day data. We use ROC-AUC and Precision-Recall AUC as the primary offline evaluation metrics. In this paper, we only report ROC-AUC and refer to it as "AUC". We observe based on past experiments that a gain in AUC on the order of +0.1\% is correlated with statistically-significant reductions in Cost-Per-Action (CPA) online. Given this is a multi-task model, we look at the performance across different objectives with an emphasis on the objectives that are used for serving.

Due to label sparsity, AUC values can be very noisy. To get more stable results, we can "average" AUC gains from evaluations on next day data over several continual training days. We define cumulative AUC gain as:

$$
\text{AUC Gain}_{cumulative} = \left(\frac{\sum_i \text{treatment\_AUC}_i}{\sum_i \text{control\_AUC}_i} -1\right) \times 100
$$

\subsection{Model Architecture}

\begin{figure}[h]
  \centering
  \includegraphics[width=\linewidth]{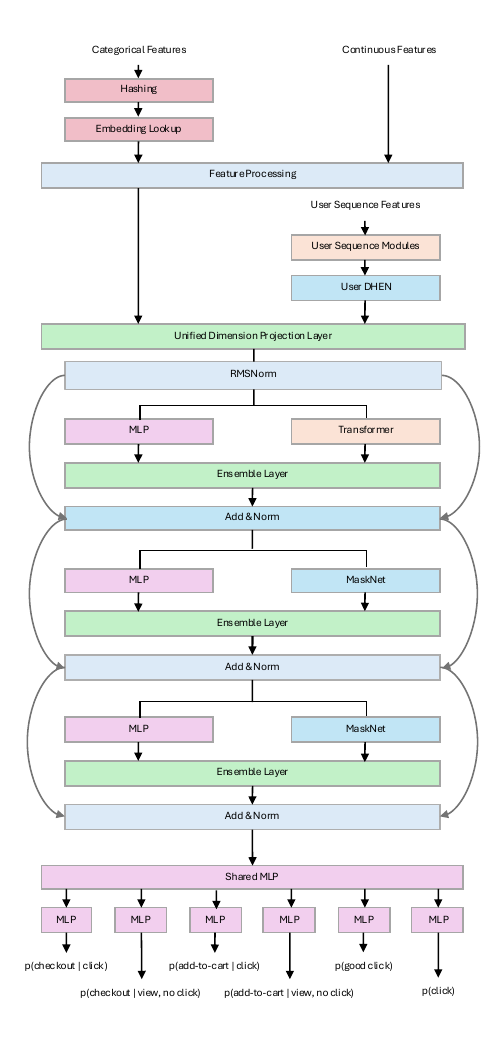}
  \caption{Sample Pinterest Ads Conversion Model Architecture}
  \label{fig:architecture}
\end{figure}

% \begin{figure*}[t]
%   \includegraphics[width=\linewidth]{figures/oCPM_Model_Architecture_Figure_Vertical.pdf}
%   \caption{Sample Pinterest Ads Conversion Model Architecture}
%   \label{fig:architecture}
% \end{figure*}

Given the sparsity of conversion labels, we use a multi-task model which trains on several related objectives simultaneously (see Table \ref{tab:head_density}). We have provided a high level architecture diagram in Figure \ref{fig:architecture}. Only objectives $p$(checkout | click) and $p$(checkout | view, no click) are served from this model. Input features undergo feature processing, where categorical features are hashed and mapped to embedding vectors, continuous features undergo min-max normalization, and pre-trained embeddings undergo batch normalization. Sequence features are encoded using a Transformer encoder \cite{vaswani2017attention}. Following feature processing, a hierarchical ensemble \cite{zhang2022dhendeephierarchicalensemble} of different modules, such as Transformer \cite{vaswani2017attention}, MaskNet \cite{wang2021masknet} and MLPs, is used. This output is then fed to task-specific MLP layers for each objective, with a sigmoid activation function.
\begin{figure}[H]
  \centering
  \begin{subfigure}[t]{\linewidth}
    \centering
    \includegraphics[width=\linewidth]{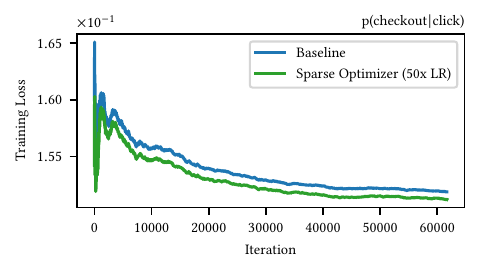}
    \caption{$p$(checkout | click)}
    \label{fig:inc_auc_gain_ccvr_a}
  \end{subfigure}

  \begin{subfigure}[t]{\linewidth}
    \centering
    \includegraphics[width=\linewidth]{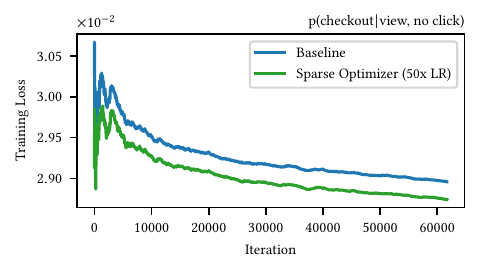}
    \caption{$p$(checkout | view, no click)}
    \label{fig:inc_auc_gain_ccvr_b}
  \end{subfigure}

  \caption{Training loss with the Sparse Optimizer}
  \label{fig:sparse_train_loss}
\end{figure}
Around 60 categorical features have their own embedding tables, most of which have very small cardinality (such as day of the week and country). A few have more the 1 million unique values including highly sparse ID features like Campaign ID, which exhibit a long tail distribution  (see Figure \ref{fig:campaign_freq_dist}). For those high cardinality features, we use binary hashing with up to 21 bits \cite{yan2021binary}. All embeddings have a fixed dimension of 32, resulting in a total of around 167 million embedding parameters out of the 298 million trainable parameters in our model.% which dominate superior to the 131 million parameters used in the feature processing/crossing and MLP layers.

%/////////////////////////////////////////////////////////////////////////////////////
\section{Experiments and Results}\label{sec:experiments}

In this section, we evaluate the efficacy of our proposed techniques against a large-scale industrial dataset. In Section \ref{sec:exp_sparse_optimizer}, we evaluate our Sparse Optimizer by comparing it to the then production model and a re-training of the production model. In Section \ref{sec:exp_multi_epoch_overfitting}, we describe the emergence of multi-epoch overfitting on our multi-task model, and evaluate the effectiveness of Frequency-Adaptive Learning Rate (FAL) on reducing multi-epoch overfitting. In Section \ref{sec:exp_larger_dataset}, we repeat our multi-epoch experiments on a larger, disjoint dataset and compare FAL to an existing multi-epoch overfitting approach, embedding re-initialization or MEDA \cite{fan2024multi}. In Section \ref{exp:continual}, we evaluate our multi-epoch models over several days of continual training.   % We are looking to answer the following questions:
% \begin{enumerate}
%     \item Does the use of a higher layer-wise learning rate for embedding tables (Sparse Optimizer), result in faster convergence and increased offline performance?
%     \item Do we exhibit multi-epoch overfitting? \begin{enumerate}
%         \item Does the use of multiple objectives via a multi-task model change the behaviour of multi-epoch overfitting?
%         \item Does the Frequency Adaptive Learning Rate (FAL) technique help mitigate multi-epoch overfitting and lead to increased offline performance?
%     \item Does seasonality and dataset size aff
%     \end{enumerate}
    
%     \item Does the Frequency Adaptive Learning Rate (FAL) technique help mitigate multi-epoch overfitting and lead to increased offline performance?
%     \item How does FAL compare with a state-of-the-art multi-epoch overfitting technique, embedding re-initialization (MEDA)?
% \end{enumerate}

\subsection{Sparse Optimizer}\label{sec:exp_sparse_optimizer}

In Figure \ref{fig:sparse_train_loss}, we show the training loss with and without the Sparse Optimizer on our two serving objectives. The Sparse Optimizer is tuned to have a 50x larger embedding table learning rate relative to the rest of the model. The Sparse Optimizer results in significantly faster convergence, as seen by a lower training loss at all points during training. At any point after the first 10000 iterations, the Sparse Optimizer achieves at least 0.41\% (0.0012) lower training loss for $p$(checkout | click) and 0.69\% (0.00026) lower training loss for $p$(checkout | view, no click).

\begin{figure}[H]
  \centering
  \includegraphics[width=\linewidth]{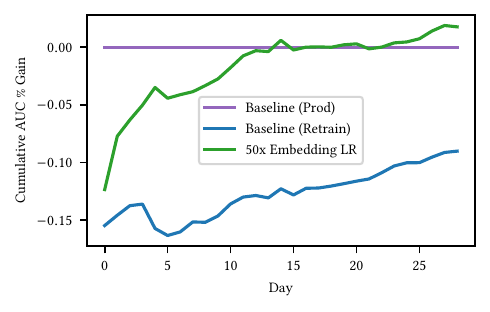}
  % \includesvg{test_plot.svg}
  % \include{test_plot}
  \caption{Cumulative offline AUC of Sparse Optimizer on $p$(checkout | view, no click) after several days of continual training.}
  \label{fig:sparse_cum_auc}
\end{figure}

In Figure \ref{fig:sparse_cum_auc}, we show the cumulative $p$(checkout | view, no click) AUC gain of the Sparse Optimizer compared to the then production model, which has an advantage of 4 additional months of continual training. We also compare to a retraining of the production model on the same training days as the Sparse Optimizer. Compared to the retrain baseline, the Sparse Optimizer results in a significantly higher cumulative AUC of 0.10\%. Moreover, our Sparse Optimizer is able to catch up with and surpass the production model 
by 0.017\% despite having a 4 month disadvantage in continual training.

% {\bf (1) We see that the use of a higher layer-wise learning rate for embedding tables (Sparse Optimizer), results in both significantly faster convergence and higher offline performance.}

\subsection{Multi-Epoch Overfitting}\label{sec:exp_multi_epoch_overfitting}

After applying the Sparse Optimizer, we batch trained our model on a dataset from May to Aug 2024, spanning 110 days. We train three models: a 1-epoch baseline, a 2-epoch baseline, and a 2-epoch treatment model using FAL.

\begin{figure}[h]
    \centering
    \includegraphics[width=\linewidth]{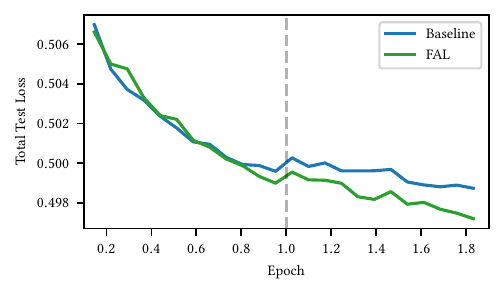}
    \caption{Total test loss progression during 2-epoch training. The training set spans May - Aug 2024.}
    \label{fig:total_loss_a}
\end{figure}

In Figure \ref{fig:total_loss_a}, we show the total test loss evolution during training, using next day data as our test set. The baseline exhibits a small increase of 0.13\% (6.82e-4) in loss at the epoch boundary similar to that observed in multi-epoch overfitting. Unlike as observed in \cite{zhang2022towards}, the increase in loss is insignificant, with the baseline recovering and ending the second epoch with a lower test loss by 1.38\% (6.92e-4) compared to the first. We hypothesize that a combination of high feature diversity, strong hashing, and increased training data through multiple label types in  multi-task learning allows the model to be more resilient to collapse.

\begin{figure*}[h]
  \centering
  \begin{tabular}{ccc}
    \begin{subfigure}[b]{0.3\linewidth}
      \includegraphics[width=\linewidth]{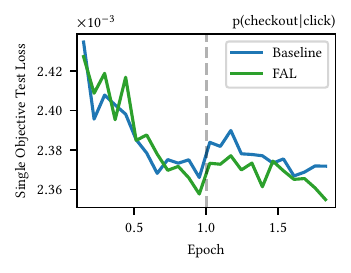}
      \caption{checkout | click}
    \end{subfigure} &
    \begin{subfigure}[b]{0.3\linewidth}
      \includegraphics[width=\linewidth]{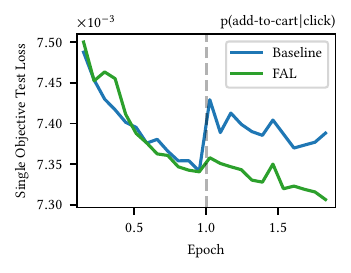}
      \caption{add-to-cart | click}
    \end{subfigure} &
    \begin{subfigure}[b]{0.3\linewidth} 
      \includegraphics[width=\linewidth]{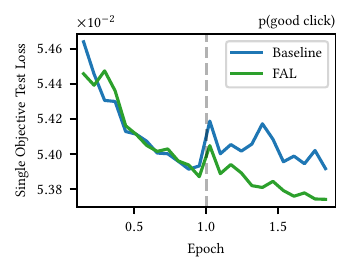}
      \caption{good click}
    \end{subfigure} \\
    \begin{subfigure}[b]{0.3\linewidth}
      \includegraphics[width=\linewidth]{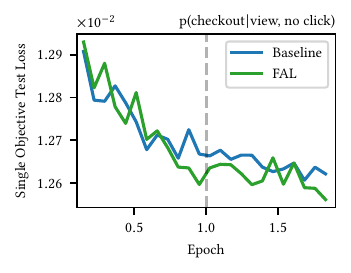}
      \caption{checkout | view, no click}
    \end{subfigure} &
    \begin{subfigure}[b]{0.3\linewidth}
      \includegraphics[width=\linewidth]{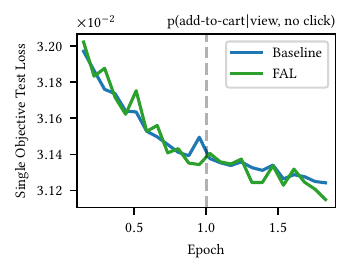}
      \caption{add-to-cart | view, no click}
    \end{subfigure} &
    \begin{subfigure}[b]{0.3\linewidth}
      \includegraphics[width=\linewidth]{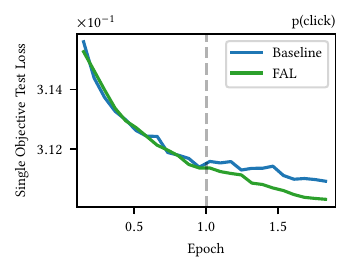}
      \caption{click}
    \end{subfigure} \\
  \end{tabular}
  \caption{Individual test loss progression for each objective in 2-epoch training. The training set spans May - Aug 2024.}
  \label{fig:head_losses_a}
\end{figure*}

\begin{figure*}[h]
  \centering
  \begin{tabular}{ccc}
    \begin{subfigure}[b]{0.3\linewidth}
      \includegraphics[width=\linewidth]{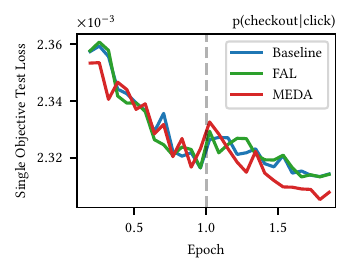}
      \caption{$p$(checkout | click)}
    \end{subfigure} &
    \begin{subfigure}[b]{0.3\linewidth}
      \includegraphics[width=\linewidth]{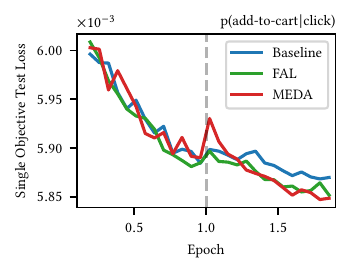}
      \caption{$p$(add-to-cart | click)}
    \end{subfigure} &
    \begin{subfigure}[b]{0.3\linewidth} 
      \includegraphics[width=\linewidth]{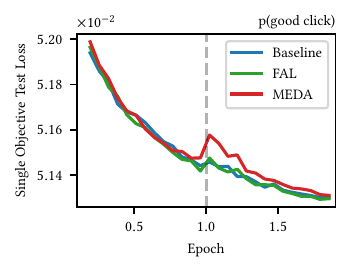}
      \caption{$p$(good click)}
    \end{subfigure} \\
    \begin{subfigure}[b]{0.3\linewidth}
      \includegraphics[width=\linewidth]{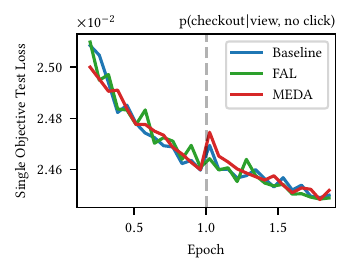}
      \caption{$p$(checkout | view, no click)}
    \end{subfigure} &
    \begin{subfigure}[b]{0.3\linewidth}
      \includegraphics[width=\linewidth]{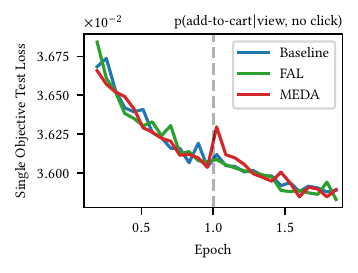}
      \caption{$p$(add-to-cart | view, no click)}
    \end{subfigure} &
    \begin{subfigure}[b]{0.3\linewidth}
      \includegraphics[width=\linewidth]{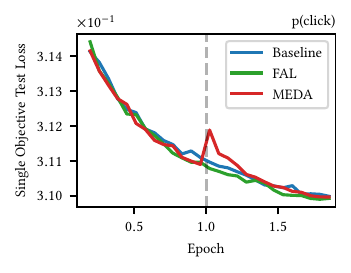}
      \caption{$p$(click)}
    \end{subfigure} \\
  \end{tabular}
  \caption{Individual test loss progression for each objective in 2-epoch training. The training set spans Aug - Dec 2024.}
  \label{fig:head_losses_b}
\end{figure*}

Total loss is a linear combination of individual losses per objective. In Figures \ref{fig:head_losses_a} and \ref{fig:head_losses_b}, we show the test loss evolution of each objective during training, using next day data as our test set. The baseline exhibits varying amounts of multi-epoch overfitting depending on the objective. In particular, the $p$(add-to-cart | click) and $p$(checkout | click) objectives exhibit a larger loss increase than their sibling objectives $p$(add-to-cart | view, no click) and $p$(checkout | view, no click). For example, $p$(add-to-cart | click) has a 1.19\% (8.80e-5) increase in loss at the epoch boundary, whereas $p$(add-to-cart | view, no click) decreases in loss by 0.37\% (1.11e-4). We observe that this is related to the relative label density of the objectives; as seen in Table \ref{tab:head_density}, clicks are rarer than views, resulting in a smaller effective training set which is more prone to overfitting. The same argument can be made with our click and sparser good click objectives.

\subsubsection{Frequency-Adaptive Learning Rate (FAL)} 

For $p$(checkout | click) and $p$(add-to-cart | click), FAL noticeably reduces the jumps in loss at the epoch boundary from +0.75\% $\to$ +0.65\% (1.78e-05 $\to$ 1.56e-05) and  +1.19\% $\to$ +0.23\% (8.80e-05 $\to$ 1.73e-05) respectively compared to the 2-epoch baseline. Moreover, it achieves better performance at the end of batch training, with the same objectives achieving 0.47\% (1.11e-05) and 0.83\% (6.13e-05) lower loss respectively compared to the baseline.

\subsection{Larger dataset, Seasonality, MEDA} \label{sec:exp_larger_dataset}

In social media platforms like Pinterest, we observe fluctuations in dataset size due to seasonality. As such, we conducted a second round of evaluation of 2-epoch training for data from Aug to Dec 2024. We surprisingly observe less overfitting during this round, a result we attribute to a larger dataset (25\% larger than the previous one). With this observation, we also evaluate a new technique from literature, embedding re-initialization or MEDA \cite{fan2024multi}, to compare against our previous datapoints.

\begin{figure}[h]
    \centering
    \includegraphics[width=\linewidth]{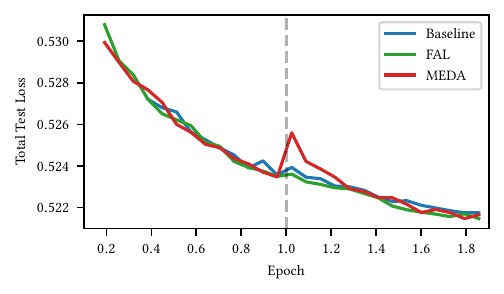}
    \caption{Total test loss progression during 2-epoch training. The training set spans Aug - Dec 2024.}
    \label{fig:total_loss_b}
\end{figure}

On total test loss, as seen in Figure \ref{fig:total_loss_b}, the baseline exhibits a much smaller increase in test loss of 0.07\% (3.66e-4) in comparison to the 0.13\% (6.82e-4) jump in our May-Aug dataset. We observe a similar pattern for individual objectives, with significantly less multi-epoch overfitting overall on the baseline. For example, the $p$(add-to-cart | click) objective displays a much smaller increase in test loss at 0.24\% (1.41e-5) compared to 1.19\% (8.80e-5)  in May-Aug. In all objectives except $p$(checkout | click) and $p$(add-to-cart | click), we exhibit negligible overfitting, with a near monotonically decreasing loss curve. In these objectives, all models end the second epoch with similar performance.

In our second sparsest objective, $p$(add-to-cart | click), both MEDA and FAL outperform the baseline in loss at the end of the second epoch by a similar amount: 0.36\% (1.81e-5) and 0.31\% (2.12-e5) respectively. In our sparsest objective, $p$(checkout | click), MEDA outperforms both FAL and the baseline in loss at the end of the second epoch by 0.27\% (6.33-e6) and 0.28\% (6.47e-6) respectively. Of note, FAL fails to alleviate multi-epoch overfitting in this objective and achieves similar performance to the baseline.

\subsection{Continual Learning} \label{exp:continual}

\begin{figure}[h]
  \centering
  \begin{subfigure}[t]{\linewidth}
    \centering
    \includegraphics[width=\linewidth]{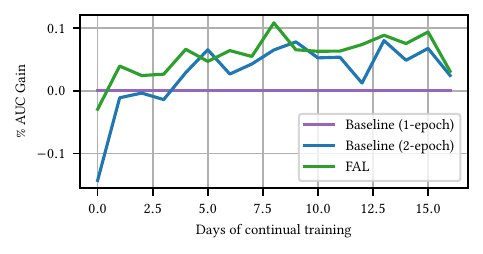}
    \caption{May-Aug 2024 Dataset}
    \label{fig:inc_auc_gain_ccvr_a}
  \end{subfigure}

  \begin{subfigure}[t]{\linewidth}
    \centering
    \includegraphics[width=\linewidth]{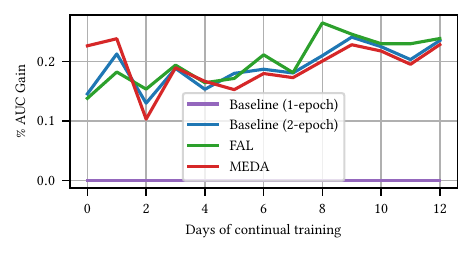}
    \caption{Aug-Dec 2024 Dataset}
    \label{fig:inc_auc_gain_ccvr_b}
  \end{subfigure}

  \caption{AUC Gain on $p$(checkout | click) head over several days of continual training for different datasets.}
  \label{fig:inc_auc_gain_ccvr}
\end{figure}

After batch training, our model undergoes continual daily trainings on fresh data. In Figure \ref{fig:inc_auc_gain_ccvr}, we evaluate the offline AUC of our models on $p$(checkout | click) over several continual trainings using next day data. 

In the May-Aug Dataset, FAL performs best, with a cumulative AUC gain of 0.07\%. Although our 2-epoch baseline initially underperforms the 1-epoch baseline due to overfitting, it achieves similar performance to FAL after 8 days of continual training.

In the Aug-Dec Dataset, MEDA initially performs best. After two days of continual training, all three models perform similarly. In comparison to the May-Aug Dataset, the gap between 2-epoch and 1-epoch models is larger, with FAL achieving the highest cumulative AUC gain at 0.20\% compared to MEDA's 0.19\% and the 2-epoch baseline's 0.18\%.

% https://reports.pinadmin.com/execution/M10N/Ranking/oCPM/MLEnv%20Offline%20Comparison/2025-02-10%2002:52:07

\section{Discussion}

\subsection{Sparse Optimizer}

Through the use of a higher layer-specific learning rate for embedding tables (Sparse Optimizer), we see compelling evidence that our models achieve both faster convergence and higher offline performance. We believe that extending layer-specific learning rates can extend beyond embedding table training. Modern recommendation systems combine a variety of feature processing modules which may converge at different speeds depending on module complexity and gradient sparsity levels. For example, certain parameters may be trained less than others due to missing features. Or, certain layers may receive fewer informative parameter updates due to positive label sparsity, as is the case with our multi-task model. We believe that carefully extending adaptive gradient methods like Adagrad \cite{JMLR:v12:duchi11a} in the context of multi-epoch overfitting can lead to fruitful results.

\subsection{Multi-Epoch Training and Frequency-Adaptive Learning Rate (FAL)}

With training and evaluation over two disjoint datasets, we show that the baseline model exhibits multi-epoch overfitting. By slicing our losses by objective in our multi-task learning setup, we showcase how multi-epoch overfitting differs in severity for each objective, dependent on its label density. A higher label density, like that observed with the $p$(click) and $p$(add-to-cart | view, no click) heads results in lower overfitting. Similarly, a larger training set size, which is observed in the Aug-Dec dataset, appears to contribute to lower overfitting across all objectives.

The use of a frequency-adaptive learning rate (FAL) aids in reducing multi-epoch overfitting. Compared to an existing method in literature, embedding re-initialization or MEDA, FAL achieves comparable performance on all objectives except the sparsest,  $p$(checkout | click). It also achieves the highest offline performance after several days of continual training.

FAL underperforms MEDA in multi-epoch overfitting. However, as a next step, we believe that its core ideas can be used to enhance existing multi-epoch overfitting methods. One disadvantage with MEDA and existing multi-epoch overfitting methods like regularization, is that it treats each embedding row equally, independent of its frequency or sparsity. By selectively lowering learning rate for infrequent rows, FAL helps alleviate multi-epoch overfitting without significantly affecting the performance of frequent rows. Unlike MEDA, this allows test losses to be approximately decreasing, allowing for the possibility of mid-epoch early stopping. We believe that a frequency-aware extension of MEDA or regularization, which reduces  emphasis on frequent rows may achieve the best of both worlds.

Although both FAL and MEDA have reduced overfitting compared our 2-epoch baseline, they achieve similar performance after several days of continual training on fresh data. In the context of an industrial batch $\to$ continual training setup, this suggests that solutions to multi-epoch overfitting may not be necessary, especially if the amount of overfitting is small. We imagine that these solutions are better suited for a multi-epoch online learning setup, in which the use of fresh data to correct overfitting before serving may not be practical, and overfitting may be more pronounced due to smaller dataset sizes. 

Both of our approaches, Sparse Optimizer and FAL, cross our internal offline cumulative AUC threshold of +0.10\% to qualify for online experiments. The Sparse Optimizer has already been launched to production models.

%//////////////////////////////////////////////
\section{Conclusion} \label{sec:discussion}

In this work, we share key learnings from the development of embedding table optimization and multi-epoch training in Pinterest Ads Conversion models. We present our use of the Sparse Optimizer, which assigns a higher layer-specific learning rate to embedding tables to circumvent the slow convergence from gradient sparsity. We show that our Sparse Optimizer achieves faster convergence and increased offline model performance. Additionally, we showcase multi-epoch overfitting on a multi-task model and how its severity varies across objectives, dependent on their label densities. We outline a novel approach to mitigate multi-epoch overfitting, called frequency-adaptive learning rate (FAL), which aggressively reduces learning rate for infrequent rows. In comparison to an existing approach in literature, embedding re-initialization or MEDA, our approach achieves similar performance in all but our sparsest objective. We validate the effectiveness of MEDA and suggest additional approaches which may combine the advantages of both FAL and MEDA. Lastly, we discuss the performance convergence of all our multi-epoch models after several days of continual training, showcasing how treating multi-epoch overfitting may be unnecessary as long as fresh data is available and overfitting is not too severe. 

\bibliographystyle{ACM-Reference-Format}
\bibliography{large_embed_bib}

\end{document}